# Deep Learning for Spatio-Temporal Modeling: Dynamic Traffic Flows and High Frequency Trading

Matthew F. Dixon[*]
Stuart School of Business
Illinois Institute of Technology

Nicholas G. Polson[†]
Booth School of Business
University of Chicago

and

Vadim O. Sokolov[‡]
Systems Engineering and Operations Research
George Mason University

May 7, 2018

**Abstract**

Deep learning applies hierarchical layers of hidden variables to construct nonlinear high dimensional predictors. Our goal is to develop and train deep learning architectures for spatio-temporal modeling. Training a deep architecture is achieved by stochastic gradient descent (SGD) and drop-out (DO) for parameter regularization with a goal of minimizing out-of-sample predictive mean squared error. To illustrate our methodology, we first predict the sharp discontinuities in traffic flow data, and secondly, we develop a classification rule to predict short-term futures market prices using order book depth. Finally, we conclude with directions for future research.

---

[*]Matthew Dixon is an Assistant Professor in the Stuart Business School, Illinois Institute of Technology. E-mail: mdixon7@stuart.iit.edu.
[†]Nicholas Polson is Professor of Econometrics and Statistics at ChicagoBooth, University of Chicago. E-mail: ngp@chicagobooth.edu.
[‡]Vadim Sokolov is an Assistant Professor in the Department of Systems Engineering and Operations Research, George Mason University. E-mail: vsokolov@gmu.edu.

# 1   Introduction

Predicting spatio-temporal flows is a challenging problem as dynamic spatio-temporal data possess underlying complex interactions and nonlinearities. Deep learning applies layers of hierarchical hidden variables to capture these interactions and nonlinearities. The theoretical roots lie in the Kolmogorov-Arnold representation theorem (Arnold, 1957; Kolmogorov, 1957) of multivariate functions, which states that any continuous multivariate function can be expressed as a superposition of continuous univariate semi-affine functions. This remarkable result has direct consequences for statistical modeling as a non-parametric pattern matching algorithm. Deep learning relies on pattern matching via its layers of univariate semi-affine functions and can be applied to both regression and classification problems. Deep learners provide a nonlinear predictor in complex settings where the input space can be very high dimensional. The deep learning paradigm for data analysis is therefore algorithmic rather than probabilistic, see (Breiman, 2001).

Traditional statistical modeling approaches use a data generating process, generally motivated by physical laws or constraints. For example, spatio-temporal flow modeling uses physical models to describe the evolution of flows (Cressie & Wikle, 2015; Richardson, Kottas, & Sansó, 2017). The application of deep layers is central to our approach and builds on the previous work of (Higdon, 1998; Stroud, Müller, & Sansó, 2001; Wikle, Milliff, Nychka, & Berliner, 2001). The advantage of using deep layers with large amounts of training data is that nonlinearities and complex interactions can be discovered at different time scales. (Di Mauro, Vergari, Basile, Ventola, & Esposito, 2017) develop deep learning models for analysis for spatio-temportal satellite images to predict land use patterns. (McDermott & Wikle, 2017) apply Bayesian deep learning techniques (Polson, Sokolov, et al., 2017) to address the issue of uncertainty quantification for spatio-temporal image analysis. Convolutional neural networks were also applied to analysis of video by (Taylor, Fergus, LeCun, & Bregler, 2010).

Training a deep learning architecture can be performed with stochastic gradient descent (SGD) which learns the weights and offsets in an architecture between the layers. Drop-out (DO) performs variable selection (Srivastava, Hinton, Krizhevsky, Sutskever, & Salakhutdinov, 2014). Deep learning (DL) relies on having a large amount of training data together with a flexible architecture to 'match' in and out of sample performance as measured by mean error, area under the curve (AUC) or the F1 score, which is the harmonic mean of precision and recall.

To illustrate our methodology, we provide two applications (i) predicting the sharp discontinuities in short-term traffic flows that arise from the regime shift in free flow to congestion and (ii) classifying short-term price movements from a limit order book of financial futures market data. Both applications exhibit sharp regime changes which are hard to capture with traditional modeling techniques.

The rest of our paper is outlined as follows. Section 1.1 provides a statistical perspective on deep learning and outlines the training, validating and testing process required to construct a deep learner. Section 2 describes dynamic spatio-temporal modeling with deep learning. Section 3 describes the deep learning model of (Polson & Sokolov, 2016) for predicting short-term traffic flows. We provide a description of the spatio-temporal formulation of the deep learning model of (Sirignano, 2016; Dixon, 2017, 2018). We quantify the empirical gains using a deep learner, to capture discontinuities and nonlinearities in the price movements, versus the elastic net method are quantified. Finally, Section 5 concludes with directions for future research.

## 1.1 Deep Learning

Deep learning is a form of machine learning that addresses a fundamental prediction problem: Construct a nonlinear predictor, $\hat{Y}(X)$, of an output, $Y$, given a high dimensional input matrix $X = (X_1, \ldots, X_P)$. Machine learning can be simply viewed as the study and construction of an input-output map of the form

$$Y = F(X) \text{ where } X = (X_1, \ldots, X_p).$$

The output variable, $Y$, can be continuous, discrete or mixed. For example, in a classification problem, $F : X \to Y$ where $Y \in \{1, \ldots, K\}$ and $K$ is the number of categories.

A deep predictor is a particular class of multivariate functions $F(X)$ that are generated by the superposition of univariate semi-affine functions. A semi-affine function, denoted by $f^l_{W^l,b^l}$, is defined as

$$f^l_{W^l,b^l}(X) := f^l(W^l X + b^l)$$

where $f^l$ is univariate and continuous. A non-linear predictor is constructed using a sequence of layers $L$ via a composite map

$$\hat{Y}(X) := F_{W,b}(X) = \left(f^L_{W^L,b^L} \ldots \circ f^1_{W^1,b^1}\right)(X).$$

Here $W = (W^1, \ldots, W^L)$ and $b = (b^1, \ldots, b^L)$ are weight matrices and offsets respectively. Deep learners form a universal basis due to the Kolmogorov-Arnold representation theorem (Kolmogorov, 1957; Arnold, 1957).

Let $Z^l$ denote the $l$-th layer with $Z^0 = X$. The structure of a deep prediction rule can then be written as a hierarchy of $L-1$ unobserved layers, $Z^l$, given by

$$\hat{Y}(X) = f^L(Z^{L-1}),$$
$$Z^1 = f^1\left(W^1 Z^0 + b^1\right),$$
$$Z^2 = f^2\left(W^2 Z^1 + b^2\right),$$
$$\ldots$$
$$Z^{L-1} = f^{L-1}\left(W^{L-1} Z^{L-2} + b^{L-1}\right).$$

When $Y$ is numeric, the output function $f^L(X)$ is given by the semi-affine function $f^L(X) := f^L_{W^L, b^L}(X)$. When $Y$ is categorical, $f^L(X)$ is a softmax function. The activation (or link) functions, $f^l$, $1 \leq l < L$, are pre-specified where as the weight matrices $W^l \in \mathbf{R}^{N_l \times N_{l-1}}$ and offset vectors $b^l \in \mathbf{R}^{N_l}$ have to be learned from a training dataset $(X^{(i)}, Y^{(i)})_{i=1}^T$. Common choices of $f^l$ are hinge or rectified linear units, $\max(x, 0)$, and sigmoidal ($\cosh(x), \tanh(x)$) activation functions. We will later discuss stochastic gradient descent (SGD) and the proximal Newton method described in Section 5.2.

The first stage in our architecture performs variable selection is to reduce the dimensionality of the problem. We first use a vector-autoregressive model or an elastic-net model (Zou & Hastie, 2005) to perform a regularized fit to find the low dimensional parameter space. For our traffic flows data, we pre-processed with lasso and for our high frequency data with an elastic net. Multi-layer deep learning network is then used to model the nonlinear spatio-temporal patterns.

There are a number of issues in any architecture design. How many layers? How many neurons $N_l$ in each hidden layer? How to perform 'variable selection'? Many of these problems can be solved by a stochastic search technique, called dropout (Srivastava et al., 2014), which we discuss in Section 5.3. Figure 1 shows a number of commonly used structures; for example, feed-forward architectures, neural Turing machines. Once you have learned the dimensionality of the weight matrices which are non-zero, there's an implied network structure.

Starting with a generic architecture, the sparsity pattern of the weight matrices and the number of weight matrices is learned to adapt the depth and shape to the data. Different architectures can be trained and compared to determine the most suitable predictive model. Different applications have arrived at preferred architectures for data representation. For example, convolutional networks are preferred for image processing where as recurrent and long/ short term memory architectures are often used for time series prediction.

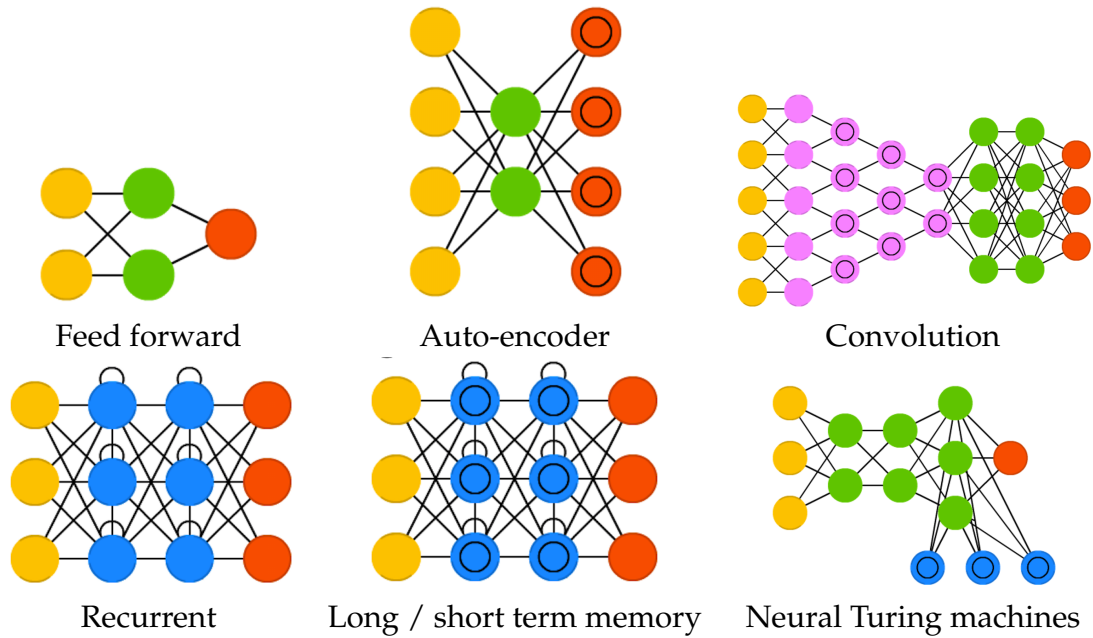

Figure 1: *Most commonly used deep learning architectures for modeling. Source:* http://www.asimovinstitute.org/neural-network-zoo.

The power of deep learners is that we can use multiple hidden layers. Think of each layer as a local filter that captures a complex interaction on some space, time scale.

## 1.2 Recurrent Layers

One approach to explicitly modeling the temporal dependencies is to use Recurrent neural networks (RNNs). RNNs use recurrent layers to capture temporal dependencies with a relatively small number of parameters. They learn temporal dynamics by mapping an input sequence to a hidden state sequence and outputs, via a recurrent layer and a feedforward layer. Let $Y_t$ denote the observed response and $Z_t$ are hidden states, then the RNN model is:

$$\text{response:} \quad \hat{Y}_t = f^2(W_z^2 Z_t + b^2),$$
$$\text{hidden states:} \quad Z_t = f^1(W^1[Z_{t-1}, X_t] + b^1),$$

where $f^1$ is an activation function such as $\tanh(x)$ and $f^2$ is either a softmax function or identity map depending on whether the response is categorical or continuous respectively. The time invariant weight matrices $W^1 = [W_z^1, W_x^1]$ and $W_z^2$ are found through training the network. $X_t$ are extremal inputs up to $k$ lags, $Z_{t-1}$ are the previous hidden states, and the hidden state is initialized to zero, $Z_{t-k} = 0$.

The main difference between RNNs and feed-forward deep learning is the use of a hidden layer with an auto-regressive component, here $W_z^1 Z_{t-1}$. It leads to a network topology in which each layer represents a time step, indexed by $t$, in order to highlight the temporal nature.

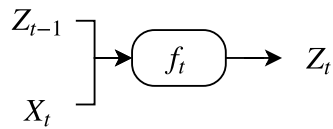

Figure 2: Hidden layer of a Recurrent Neural Network.

Additional depth can be added to create deep RNNs by stacking layers on top of each other, using the hidden state of the RNN as the input to the next layer. RNNs architectures are learned through the same mechanism described for feedforward architectures. One key difference between implementations of RNNs is that drop-out is not applied to the recurrent connections, only to the non-recurrent connections. In contrast, drop-out is applied to all connections in a feedforward architectures. Further evidence of the success of RNNs by only applying drop-out to the nonrecurrent connections is provided in (Graves, 2013).

RNNs have difficultly in learning long-term dynamics, due in part to the vanishing and exploding gradients that can result from back propagating the gradients down through the many unfolded layers of the network. A particular type of RNN, called a LSTM (Long short-term memory) network was proposed to address this issue of vanishing or exploding gradients. A memory unit used in LSTM networks allows the network to learn which previous states can be forgotten (Hochreiter & Schmidhuber, 1997; Schmidhuber & Hochreiter, 1997). For this reason, LSTMs have demonstrated much empirical success in the engineering literature (Gers, Eck, & Schmidhuber, 2001; Zheng, Xu, Zhang, & Li, 2017).

The hidden state is generated via another hidden cell state $C_t$ that allows for long term dependencies to be "remembered". Then we generate:

$$\text{Output: } Z_t = O_t \star \tanh(C_t),$$
$$K_t = \tanh(W_c^T[Z_{t-1}, X_t] + b_c),$$
$$C_t = F_t \star C_{t-1} + I_t \star K_t,$$
$$\text{State equations: } \begin{pmatrix} I_t \\ F_t \\ O_t \end{pmatrix} = \sigma(W^T[Z_{t-1}, X_t] + b),$$

where $\star$ denotes point-wise multiplication. Then, $F_t \star C_{t-1}$ introduces the long-range dependence. The states $(I_t, F_t, O_t)$ are input, forget and output states. Figure 3 shows the network architecture.

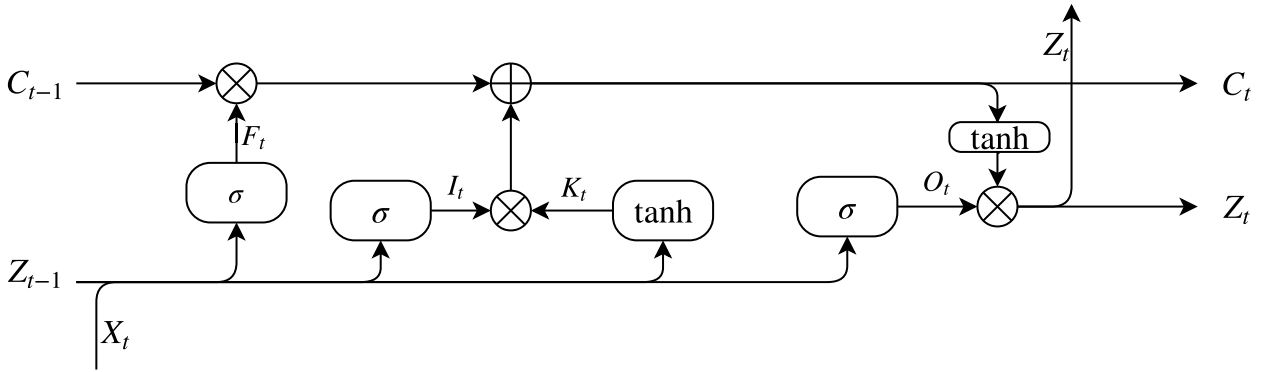

Figure 3: Hidden layer of an LSTM model. Input $(Z_{t-1}, X_t)$ and state output $(Z_t, C_t)$.

The key addition of a LSTM, compared to a RNN, is the cell state $C_t$. The information is added or removed from the memory state via gates defined via the activation function $\sigma(x)$ and point-wise multiplication $\star$. The first gate $F_t \star C_{t-1}$, called the "forget gate", selectively ignores some of the data from the previous cell state. The next gate $I_t \star K_t$, called the 'input gate', decides which values to update. Then, the new cell state is a sum of the previous cell state, passing through the forget gate selected components of the $[Z_{t-1}, X_t]$ vector. Thus, the vector $C_t$ provides a mechanism for dropping irrelevant information from the past, and adding relevant information from the current time step. The output is the result of the output gate $Z_t = O_t \star \tanh(C_t)$, which returns $\tanh$ applied to the cell state, with some entries removed. And so the forget gate is the key component for resolving the problem of vanishing and exploding gradients.

## 2 Dynamic Spatio-Temporal Modeling (DSTM)

Suppose that we have data at spatial locations $s_i, i = 1, \ldots, n$ and time locations $t_j, j = 1, \ldots, T$. Denote this by the process $Y_t = \{Y(s_i, t)\}_{i=1}^n$, where $t$ indexes the observation scale. The goal is to predict at a particular location, and forecast at a new time point given training data. Denote this quantity by $Y(s^*, t^*)$. A simple linear predictor would take the form as a "local" average of "near-by" points denoted by

$$\hat{Y}(s^*, t^*) = \sum_{i=1}^n \sum_{j=1}^T w_{ij}^* Y(s_i, t_j). \tag{1}$$

Deep learning simply uses a hierarchical layered predictor with univariate activation functions and weight matrices of different dimensions to capture the more complicated structures and relationships that exist in the evolution of the process in space and time. Typical architectures (connecting non-zero weights) include traditional RNNs, convolutional neural networks and, more recently, LSTMs. Deep Learning is an algorithmic approach rather than probabilistic in its nature, see (Breiman, 2001) for the merits of both approaches.

Rather than directly imposing a covariance structure (e.g. Gaussian process with O($n^3$) parameters (Gramacy & Polson, 2011)), a deep learner provides a flexible functional form to directly model the predictor, $\hat{Y}$. Parameter search is then achieved by regularizing a measure of fit and the optimal amount of regularization is achieved by measuring the out-of-sample bias-variance trade-off in a hold-out sample. Underlying this approach is the assumption that we have sufficient data to 'train' a predictor that captures the hidden complex interactions. See Appendix A for further discussions of training with SGD.

It is instructive to see the corresponding RNN predictor for the spatio-temporal model

above:

$$\hat{Y}_{t^*}(s) = f^2(W_z^2 Z_t + b^2),$$
$$Z_{t-T} = f^1\left(W^1[0, Y_{t-T}] + b^1\right),$$
$$Z_{t-T+1} = f^1\left(W^1[Z_{t-T}, Y_{t-T+1}] + b^1\right),$$
$$\dots$$
$$Z_t = f^1\left(W^1[Z_{t-1}, Y_t] + b^1\right).$$

where $\hat{Y}(s_i, t^*) = \hat{Y}_{t^*}(s_i)$ is the model output at location $s_i$ and time $t^*$ and each hidden state $Z_t \in \mathbb{R}^n$. Stated in this simple form, it is easy to see that RNNs are just non-parametric analogs of non-linear vector auto-regressive models.

We can also draw the analogy between filtering techniques traditionally used for spatio-temporal modeling such as Kalman filters, and RNNs. In filtering techniques, we model the relation between measured data $Y_t$ and hidden state vectors $Z_t$ using two probabilistic models, the measurement model $p(Y_t|Z_t)$ and the transition model $p(Z_{t+1}|Z_t)$. Bayes' rule is used to calculate $p(Z_t|Y_t)$. By contrast, RNNs learn a deterministic map from $Y_t$ to $Z_t$ using back-propagation.

The inclusion of an auto-regressive component in deep learners has direct consequences for modeling and input data configuration. In the feedforward architecture, the time dimension is represented implicitly - lagged input variables are embedded into the input vector. The input weight matrix $W^1$ can be scaled by a factor of $k$ and the dimension of the hidden weight matrices are increased accordingly. A feed-forward network with lagged observations permits the number of lags to vary in space - the sparsity structure of the input matrix determines which lagged variables are included in the model.

RNNs represent the time dimension explicitly - they do not require the embedding of lagged input variables in the input feature space. Instead, a single layer of $n$ units is 'un-folded' k times to represent the time dimension. Hence the dimension of the recurrence and output weight matrices $W_z^1$ and $W_z^2$ is independent of $k$. For this reason, RNNs are typically smaller and easier to train than feed-forward networks. A RNN, however, fixes the number of lags across space and time and thus does not allow for such a flexible representation of the data.

For a RNN, the number of weights in our experiments is generally under a hundred, but can increase to thousands in larger datasets from these applications. In contrast, the total number of weights in a feed-forward architectures is observed to be of the order of thousands to tens of thousands. Further discussion of the configuration of spatio-temporal deep learners is discussed in our applications.

## 3 Applications: Dynamic Traffic Flows

### 3.1 Predicting Traffic Flow Speeds

To illustrate our methodology, we use data from twenty-one loop-detectors installed on a northbound section of Interstate I-55 which span 13 miles of the highway in Chicago[1]. A loop-detector is a presence sensor that measures when a vehicle is present and generates an on/off signal. Since 2008, Argonne National Laboratory has been archiving traffic flow data every five minutes from the grid of sensors recording averaged *speed*, *flow*, and *occupancy*. Occupancy is defined as the percentage of time a point on the road is occupied by a vehicle, and flow is the number of on-off switches. Illinois uses a single loop detector setting, and speed is estimated based on the assumption of an average vehicle length.

Finding the spatio-temporal relations in the data is the predictor selection problem. Figure 4 illustrates a space-time diagram of traffic flows on the 13-mile stretch of highway I-55. You can see a clear spatio-temporal pattern in traffic congestion propagation in both downstream and upstream directions. The spatio-temporal data can be represented as

$$Y_t = x_{t+h}^t = \begin{pmatrix} x_{1,t+h} \\ \vdots \\ x_{n,t+h} \end{pmatrix},$$

$x_{t+h}^t$ is the forecast of traffic flow speeds at time $t+h$, given measurements up to time $t$. Here $n$ is the number of locations on the network (loop detectors) and $x_{i,t}$ is the cross-sectional traffic flow speed at location $i$ at time $t$. For the traffic flow model previously measured and possibly

---

[1] Traffic flow data is available from the Illinois Department of Transportation, (see Lake Michigan Interstate Gateway Alliance `http://www.travelmidwest.com/`, formally the Gary-Chicago-Milwaukee Corridor, or GCM).

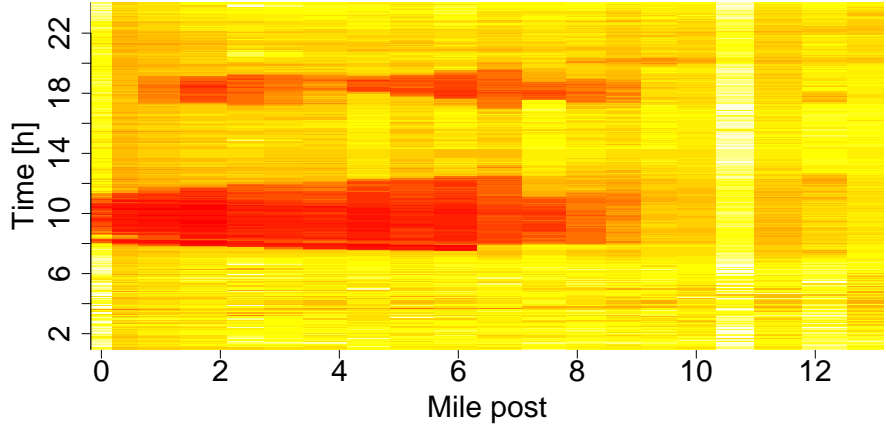

Figure 4: *A space-time diagram that shows traffic flow speed over a 13-mile stretch of I-55 in Chicago on 18 February 2009 (Wednesday). Red represents slow speed and light yellow corresponds to free flow speed. The direction of the flow is from 0 to 13.*

filtered traffic flow data given by $x^t = (x_{t-k}, \ldots, x_t)$ used as predictors

$$x = x^t = \text{vec} \begin{pmatrix} x_{1,t-k} & \ldots & x_{1,t} \\ \vdots & \vdots & \vdots \\ x_{n,t-k} & \ldots & x_{n,t} \end{pmatrix}.$$

$k$ is the number of previous measurements used to develop a forecast and `vec` is the vectorization transformation which converts the matrix into a column vector.

In our application examined later in Section 3.2, we have twenty-one road segments (i.e., $n = 21$) that span thirteen miles of a major corridor connecting Chicago's southwest suburbs to the central business district. The chosen length is consistent with several current transportation corridor management deployments (TransNet, 2016). The prediction horizon $h = 40$ is chosen as the 90th percentile of trip length in the Chicago metropolitan area, so that the model applies to most travelers. $k$ is chosen empirically so that the look-back period is 60 minutes.

Our layers are constructed as follows, $Z^0 = x$, then $Z^l$, $l = 1, \ldots, L$ is a time series "filter" given by

$$Z^l = f\left(W^l Z^{l-1} + b^l\right).$$

Here $Z^{l-1} \in \mathbb{R}^{N_{l-1}}$ denotes a vector of inputs into a layer $l$ and $N_l$ is the number of activation units (neurons) in layer $l$ and function $f$ is called an activation function.

A predictor selection problem requires estimation algorithms for finding sparse models.

Those rely on adding a penalty term to a loss function. A recent review by (Nicholson, Matteson, & Bien, 2017) considers several prominent scalar regularization terms to identify sparse vector auto-regressive models.

First we construct a hierarchical linear vector autoregressive model to identify the spatio-temporal relations in the data. We consider the problem of finding sparse matrix, $W^0$, in the following model

$$x_{t+h}^t = W^0 x^t + \epsilon_t, \quad \epsilon_t \sim N(0, V);$$

where $W^0$ is a matrix of size $n \times nk$. In our example in Section 3.2, we have $n = 21$; however, in large scale sensor networks, there are tens of thousands locations with measurements available.

The predictors selected as a result of finding the linear model are then used to build a deep learning model. To find an optimal network (structure and weights) we used the SGD method implemented in the package `H2O`. Similar methods are available in `Python`'s `Theano` (Bastien et al., 2012) or `TensorFlow` (Abadi et al., 2016) framework. We use random search to find meta parameters of the deep learning model. To illustrate our methodology, we generated $N = 10^5$ Monte Carlo samples from the following feed-forward network architecture:

$$\text{response:} \quad \hat{Y}_t = W^L Z^{L-1} + b^L,$$
$$\text{hidden states:} \quad Z^l = \tanh(W^l Z^{l-1} + b^l), \ l \in \{1, \ldots, L-1\},$$

where $L = 4$ and the network is tapered so that $W^1 \in \mathbb{R}^{150 \times 252}, W^2 \in \mathbb{R}^{100 \times 150}, W^3 \in \mathbb{R}^{50 \times 100}$ and $W^4 \in \mathbb{R}^{n \times 50}$.

**Alternative architectures** We mention in passing that other architectures are feasible for this problem, for example the vanilla RNN given in Section 2 would be configured as

$$\text{response:} \quad \hat{Y}_t = W_z^2 Z_t + b^2,$$
$$\text{hidden states:} \quad Z_{t-j} = \tanh(W^1[Z_{t-j-1}, X_{t-j}] + b^1), j \in \{k, \ldots, 0\},$$

where $W_x^1 \in \mathbb{R}^{12 \times 21}, W_z^1 \in \mathbb{R}^{12 \times 12}$ and $W_z^2 \in \mathbb{R}^{n \times 12}$ and the hidden states are initialized to zero, $Z_{t-k} = 0$. We reiterate the main difference between the configuration of spatio-temporal feed-forward networks and plain RNNs: in the feedforward architecture, the time dimension is represented implicitly - lagged input variables are embedded into the input vector and the number of input neurons is $kn$. The input weight matrix $W^1$ is scaled by a factor of $k$ and dimension of the hidden weight matrices are increased accordingly. RNNs do not require the embedding of lagged input variables in the input feature space which explains why the dimension of the recurrence and output weight matrices $W_z^1$ and $W_z^2$ are much smaller than the feed-forward network weight matrices.

**Training** To find the optimal structure of the feed forward network (number of hidden layers $L$, number of activation units in each layer $N_l$ and activation functions $f$) as well as hyper-parameters, such as $\ell_1$ regularization weight, we used a random search. Though this technique can be inefficient for large scale problems, for the sake of exploring potential structures of the networks that deliver good results and can be scaled, this is an appropriate technique for small dimensions. Stochastic gradient descent is used for training as it scales linearly with the data size. Thus the hyper-parameter search time is linear with respect to model size and data size. On a modern processor it takes about two minutes to train a deep learning network on 25,000 observations of 252 variables. Hyper-parameter tuning and model structure search requires the model to be fit $10^5$ times. Thus the total wall-time (time that elapses from start to end) was 138 days. An alternative to random search for learning the network structure for traffic forecasts was proposed in (Vlahogianni, Karlaftis, & Golias, 2005) and relies on the genetic optimization algorithm.

## 3.2 Traffic Flow on Chicago's Interstate I-55

One of the key attributes of congestion propagation on a traffic network is the spatial and temporal dependency between bottlenecks. For example, if we consider a stretch of highway and assume a bottleneck, than it is expected that the end of the queue will move from the bottleneck downstream. Sometimes, both the head and tail of the bottleneck move downstream together. Such discontinuities in traffic flow, called shock waves are well studied and can be modeled using a simple flow conservation principles. However, a similar phenomena can be observed not only between downstream and upstream locations on a highway. A similar relationship can be established between locations on city streets and highways (Horvitz, Apacible, Sarin, & Liao, 2012).

An important aspect of traffic congestion is that it can be 'decomposed' into recurrent and non-recurrent factors. For example, a typical commute time from a western suburb to Chicago's city center on Mondays is 45 minutes. However, occasionally the travel time is 10 minutes shorter or longer. Figure 5(a) shows measurements from all non-holiday Wednesdays in 2009. The solid line and band, represent the average speed and 60% confidence interval respectively. Each dot is an individual speed measurement that lies outside of 98% confidence interval. Measurements are taken every five minutes, on every Wednesday of 2009; thus, we have roughly 52 measurements for each of the five-minute intervals.

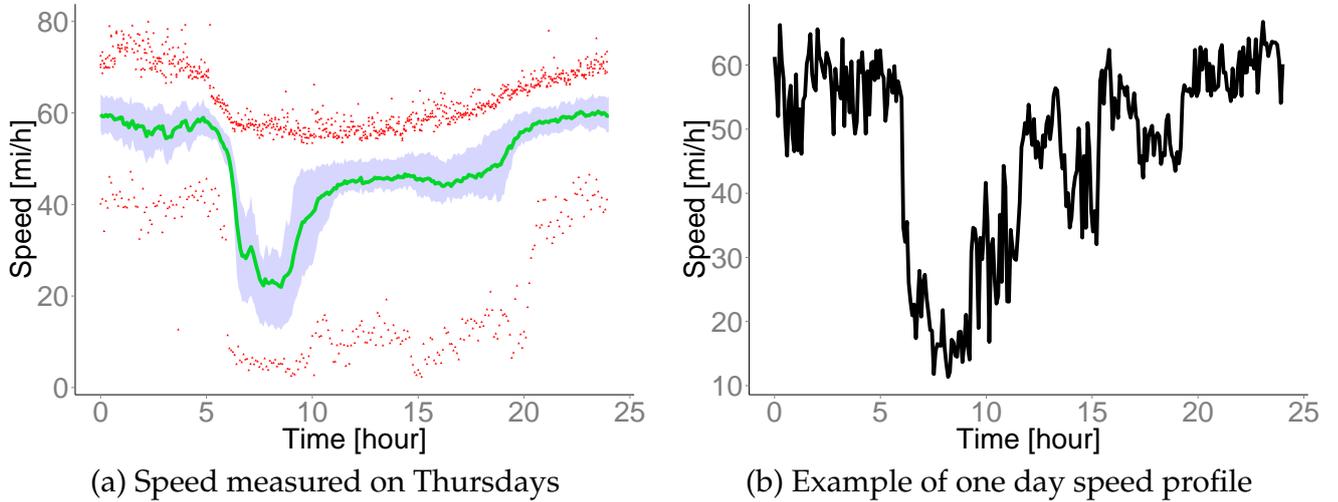

(a) Speed measured on Thursdays  (b) Example of one day speed profile

Figure 5: Recurrent speed profile. Both plots show the speed profile for a segment of interstate highway I-55. Left panel (a) shows the green line, which is the average cross-section speed for each of five minute intervals with 60% confidence interval. The red points are measurements that lie outside of 98% confidence interval. Right panel (b) shows an example of one day speed profile from May 14, 2009 (Thursday).

In many cases traffic patterns are very similar from one day to another. However, there are many days when we see surprises, both good and bad. A good surprise might happen, e.g., when schools are closed due to extremely cold weather. A bad surprise might happen due to non-recurrent traffic conditions, such as an accident or inclement weather.

Figure 6 shows the impact of non-recurrent events. In this case, the traffic speed can significantly deviate from historical averages due to the increased number of vehicles on the road or due to poor road surface conditions.

Our goal is to build a statistical model to capture the sudden regime changes from free flow to congestion and then the decline in speed to the recovery regime for both recurrent and non-recurrent traffic conditions. To this end, we compare the overall performance of our deep learning (DL) model with a sparse linear vector autoregression (VAR) model and assess the relative capability to capture these sudden regime changes. In our empirical study, we predict traffic flow speed at the location of Sensor 11, which is in the middle of the 13-mile stretch. Thus we analyze one component of the model output vector $Y$. Missing data is estimated by linear interpolation in space, i.e. the missing speed measurement $x_{i,t}$ for sensor $i$ at time $t$ will be estimated using $(x_{i-1,t} + x_{i+1,t})/2$. We exclude public holidays, weekends and days when there is a sensor network failure.

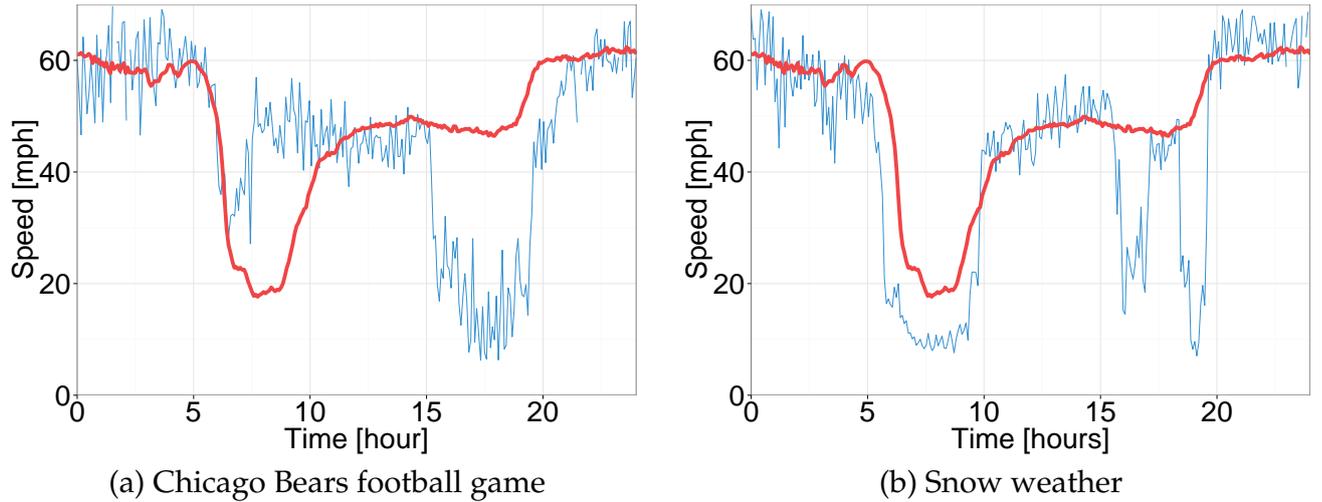

(a) Chicago Bears football game  (b) Snow weather

Figure 6: *Impact of non-recurrent events on traffic flows. Left panel (a) shows traffic flow on a day when New York Giants played at Chicago Bears on Thursday October 10, 2013. Right panel (b) shoes impact of light snow on traffic flow on I-55 near Chicago on December 11, 2013. On both panels average traffic speed is red line and speed on event day is blue line.*

Each model is combined with several data pre-filtering techniques, namely median filtering (Arce, 2005) with a window size of 8 measurements (M8) and trend filtering (Kim, Koh, Boyd, & Gorinevsky, 2009) with $\lambda = 15$ (TF15). We also test the performance of the sparse linear model, identified via regularization. We estimate the percent of variance explained by each model, and mean squared error (MSE), which measures the average of the deviations between measurements and model predictions. To train both models we choose a contiguous observation period of 90 days in 2013. We further choose another contiguous 90 day period in 2013 for testing. $R^2$ and MSE for both in-sample (IS) data and out-of-sample (OS) data are shown in Table 1.

|        | DLL   | DLM8L | DLM8  | DLTF15L | DLTF15 | VARM8L | VARTF15L |
|--------|-------|-------|-------|---------|--------|--------|----------|
| IS MSE | 13.58 | 7.7   | 10.62 | 12.55   | 12.59  | 8.47   | 15       |
| IS $R^2$ | 0.72 | 0.83 | 0.76  | 0.75    | 0.75   | 0.81   | 0.7      |
| OS MSE | 13.9  | 8.0   | 9.5   | 11.17   | 12.34  | 8.78   | 15.35    |
| OS $R^2$ | 0.75 | 0.85 | 0.82  | 0.81    | 0.79   | 0.83   | 0.74     |

Table 1: This table compares the in-sample and out-of-sample metrics across different models. The abbreviations for column headers are: DL = deep learning, VAR = linear model, M8 = media filter preprocessing, TF15 = trend filter preprocessing and L = sparse estimator (lasso). The abbreviations for row headers are: IS = in-sample, MSE = mean squared error and OS = out-of-sample.

Comparing the out-of-sample performance, we observe that sparse deep learning combined with the median filter pre-processing (DLM8L) is the most favorable. Figure 7 compares the performance of both vector auto-regressive and deep learning models for a normal day, a special event day (Chicago Bears football game) and a poor weather day (snow day). The performance of each model is not uniform throughout the day - the absolute value of the residuals (red circles) against the measured data (black line) are shown. The highest residuals are observed when the traffic flow regime changes from one to another. On a normal day large errors are observed at around 6am, when the regime changes from free flow to congestion and at around 10am, right before it resumes free flow.

Comparing the model out-of-sample performance, we observe that sparse deep learning combined with the median filter pre-processing (DLM8L) is the most favorable. Both deep learning (DL) and vector auto-regressive (VAR) models accurately predict the morning rush hour congestion on a normal day. However, the vector auto-regressive model mis-predicts congestion during evening rush hour. At the same time, the deep learning model does predict breakdown accurately but miss-estimates the time of recovery. Both deep learning and linear model outperform naive forecasting when combined with data pre-processing. However, when unfiltered data is used to fit deep learning combined with a sparse linear estimator (DLL) model, their predictive power degrades and were out-performed by a naive forecast. These results highlight the importance of using filtered data to develop forecasts.

# 4 Applications: High Frequency Trading

Modern financial markets facilitate the electronic trading of financial instruments through an instantaneous double auction. At each point in time, the market demand and the supply can be represented by an electronic limit order book, a cross section of orders to execute at various price levels away from the market price.

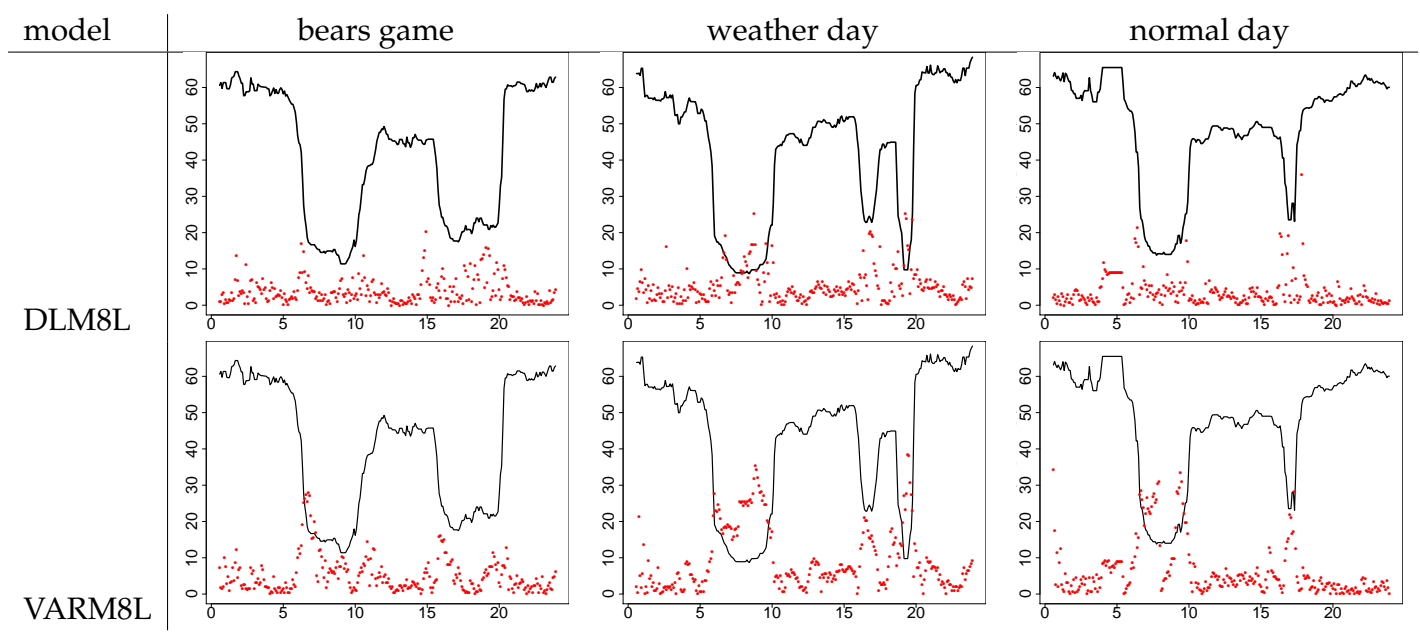

Figure 7: This figure shows the residuals of the model forecasts against time of day. In all plots the **black** solid line is the measured data (cross-section speed), the **red** dots are absolute values of residuals from our models' forty minute forecast. The first column compares models for data recorded on Thursday October 10, 2013 - the day when the Chicago Bears team played the New York Giants. The game started at 7pm and caused an unusual pattern of congestion, starting at around 4pm. The second column compares models for data recorded on Wednesday December 11, 2013, a day of light snow. The snow led to heavier congestion during both, the morning and evening rush hours. The third column compares models for data recorded on Monday October 7, 2013 - a 'normal day' on which no special events, accidents or inclined weather conditions occurred.

The market price is closely linked to its liquidity - that is the immediacy in which the instrument can be converted into cash. The liquidity of markets are characterized by their depth, the total quantity of quoted buy and sell orders about the market price. Liquid markets are attractive to market participants as they permit the near instantaneous execution of large volume trades at the best available price, with marginal price impact. A participant enters into a trade by submitting an order to a queue and either waits up to a few milliseconds for the order to be filled or cancels the order. This type of trading adds liquidity and is said to be 'making a market', a primary function of high frequency trading firms. A participant willing to pay a premium to trade at the best price can by pass the queue and is said to be 'market taking'. The liquidity of the market evolves in response to trading activity (Bloomfield, O'Hara, & Saar, 2005); At any point in time, the amount of liquidity in the market can be characterized by the cross-section of book depths. The price levels closest to the market price define the 'inside market' and is the most actively traded.

The field of microstructure research (Parlour, 1998; Cao, Hansch, & Wang, 2009; Cont, Kukanov, & Stoikov, 2014) has established a causational relationship between the depth of the inside market and the market price through temporal models of order flow imbalance. There is growing evidence that the study of microstructure is critical to studier longer term relations and even cross-market effects (Dobrislav & Schaumburg, 2016).

Recently microstructure researchers have looked beyond the inside market to predict the price movement. Most notably, (Kozhan & Salmon, 2012), use a series of independent regressions to forecast each price level. The link between dynamic spatio-temporal models is demonstrated here. We build on previous machine learning algorithms for futures price predictions with high frequency data (Sirignano, 2016; Dixon, 2017, 2018).

## 4.1 Predicting High Frequency Futures Prices

Our dataset is an archived Chicago Mercantile Exchange (CME) FIX format message feed captured from August 1, 2016 to August 31, 2016. This message feed records all transactions in the E-mini S&P 500 (ES) between the times of 12:00pm and 22:00 UTC. We extract details of each limit order book update, including the nano-second resolution time-stamp, the quoted price and depth for each limit order book level.

Figure 8 illustrates the intuition behind a typical mechanism resulting in mid-price movement. We restrict consideration to the top five levels of the ES futures limit order book, even though there are updates provided for ten levels. The chart on the left represents the state of the limit order book prior to the arrival of a sell aggressor. The x-axis represents the price levels and the y-axis represents the depth of book at each price level. Red denotes bid orders and blue denotes ask orders. The highest bid price ('best bid') is quoted at $2175.75 with a depth of 103 contracts. The second highest bid is quoted at $2175.5 with a depth of 177 contracts. The lowest ask ('best ask' or 'best offer') is quoted at $2176 with 82 contracts and the second lowest ask is quoted at $2176.25 with 162 contracts.

The chart on the right shows the book update after a market crossing limit order ('aggressor') to sell 103 contracts at $2175.75. The aggressor is sufficiently large to match all of the best bids. Once matched, the limit order is updated with a lower best bid of $2175.5. The gap between the best ask and best bid would widen if it weren't for the arrival of 23 new contracts offered at a lower ask price of $2175.75. The net effect is a full down-tick of the mid-price.

Table 2 shows the corresponding spatio-temporal representation of the limit order book before and after the arrival of the sell aggressor. The response is mid-price movement, in units of ticks, over the subsequent interval. $p_{i,t}^b$ and $d_{i,t}^b$ denote the level $i$ quoted bid price and depth of the limit order book at time $t$. $p_{i,t}^a$ and $d_{i,t}^a$ denote the corresponding level $i$ quoted ask price and depth. Level $i = 1$ corresponds to the best ask and bid prices. The mid-price at time $t$ is denoted by

$$p_t = \frac{p_{1,t}^a + p_{1,t}^b}{2}. \qquad (2)$$

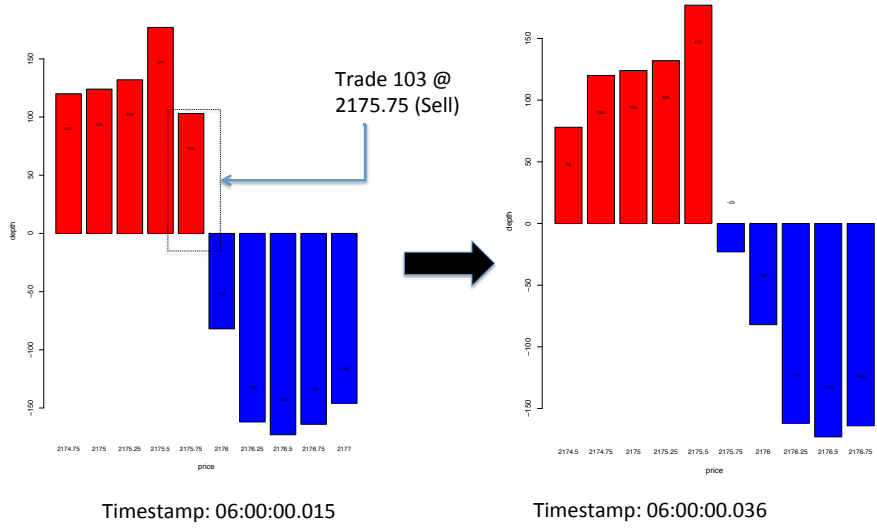

Figure 8: *This figure illustrates a typical mechanism resulting in mid-price movement. The charts on the left and right respectively show the limit order book before and after the arrival of a large sell aggressor. The aggressor is sufficiently large to match all of the best bids. Once matched, the limit order is updated with a lower best bid of $2175.5. The gap between the best ask and best bid would widen if it weren't for the arrival of 23 new contracts offered at a lower ask price of $2175.75. The net effect is a full down-tick of the mid-price.*

This mid-price can evolve in minimum increments of half a tick but almost always is observed to move at increments of a tick over time intervals of a milli-second or less.

| Timestamp | $p_{1,t}^b$ | $p_{2,t}^b$ | ... | $d_{1,t}^b$ | $d_{2,t}^b$ | ... | $p_{1,t}^a$ | $p_{2,t}^a$ | ... | $d_{1,t}^a$ | $d_{2,t}^a$ | ... | Response |
|---|---|---|---|---|---|---|---|---|---|---|---|---|---|
| 06:00:00.015 | 2175.75 | 2175.5 | ... | 103 | 177 | ... | 2176 | 2176.25 | ... | 82 | 162 | ... | -1 |
| 06:00:00.036 | 2175.5 | 2175.25 | ... | 177 | 132 | ... | 2175.75 | 2176 | ... | 23 | 82 | ... | 0 |

Table 2: *This table shows the corresponding spatio-temporal representation of the limit order book before and after the arrival of the sell aggressor listed in Figure 8. The response is the mid-price movement over the subsequent interval, in units of ticks. $p_{i,t}^b$ and $d_{i,t}^b$ denote the level $i$ quoted bid price and depth of the limit order book at time $t$. $p_{i,t}^a$ and $d_{i,t}^a$ denote the corresponding level $i$ quoted ask price and depth.*

The result of categorizing (a.k.a. labeling) the data leads to a class imbalance problem as approximately 99.9% of the observations have a zero response. This imbalance can be partially resolved by under-sampling the data at regular intervals, an approach referred to as 'clocking'. However, the imbalance is still too severe for robust classification and clocking the data set reduces the predictive power of the models. To construct a 'balanced' training set, the minority classes are oversampled with replacement and the majority class is undersampled without replacement. The resulting balanced training set has 298062 observations for ESU6.

Our model of mid-price impact is described as follows. The response is

$$Y_t = \Delta p_{t+h}^t, \tag{3}$$

where $\Delta p_{t+h}^t$ is the forecast of discrete mid-price changes from time $t$ to $t + h$, given measurement of the predictors up to time $t$. When the historical data is clocked, $h$ corresponds to the undersampling frequency. When the unclocked data is used, $h$ denotes the inter-event arrival time and will vary with trading activity. Without loss of generality, we shall set $h = 1$, and demonstrate the application of deep learning to the next period mid-price change. Our price impact model $\hat{Y}(x)$ uses relative market depth as the predictors

$$x = x^t = \text{vec} \begin{pmatrix} x_{1,t-k} & \cdots & x_{1,t} \\ \vdots & & \vdots \\ x_{n,t-k} & \cdots & x_{n,t} \end{pmatrix} \tag{4}$$

where $n$ is the number of quoted price levels, $k$ is the number of lagged observations, and $x_{i,t} \in [0,1]$ is the relative depth, representing liquidity imbalance, at quote level $i$

$$x_{i,t} = \frac{d_{i,t}^b}{d_{i,t}^a + d_{i,t}^b}. \tag{5}$$

This price impact model captures the spatio-temporal relationship between mid-price movement and the liquidity imbalance across all levels of the limit order book. The CME futures data gives $n = 10$ quote levels on either side of the market, although other exchanges such as the NYSE may release quotes for hundreds of price levels.

Figure 9 shows a time-space diagram of the limit order book. The contemporaneous depth imbalances at each price level in the limit order book polarize prior to each price movement. The x axis shows the prices of each book level and the y-axis shows the timestamp of the limit order book at 1 second snapshots over a fifteen minute period from bottom to top. In a liquid market these prices move in unison, separated by a small increment referred to as a 'tick'. However, in periods of less liquidity, temporary perturbations in the price increments near the market price may exist and the price levels temporarily fall out-of-lock step with each other.

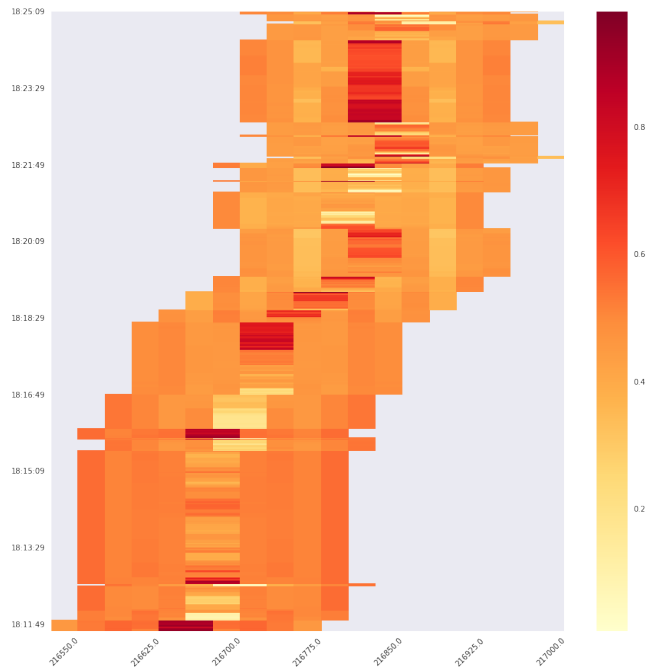

Figure 9: *A space-time diagram showing the limit order book. The contemporaneous depths imbalances at each price level, $x_{i,t}$, are represented by the color scale: red denotes a high value of the depth imbalance and yellow the converse. The limit order book are observed to polarize prior to a price movement.*

The color scale represents the liquidity imbalance relative to each price level. Red presents an excess of demand to supply and yellow the converse. With this spatio-temporal representation we may gain an appreciation as to why practitioners commonly refer to this imbalance as 'book pressure'. The book pressure at the inside market, 'the inside market pressure', is the strongest predictor of market price movement. More often than not, an upward price movement follows the cumulation of the inside market book pressure and, conversely, a downward price movement follows a depreciation of the inside market book pressure. Sometimes the full saturation or desaturation of the inside market pressure does not result in a price movement. This observation is consistent with various studies in different markets such as (Kozhan & Salmon, 2012). In using the cross-section of relative depths in the spatio-temporal model, rather than just the inside market, the deep learner is able to find relationships which lead to improved price impact forecasts, especially at the short time scales necessary for high frequency trading.

## 4.2 Deep Learner

Each categorical response $Y$ is represented as a 1-of-K indicator vector, with all elements equal to zero except the element corresponding to the correct class $k$. For example, if $K = 3$ and the correct class is $1$ then $Y$ is represented as $[1, 0, 0]$.

We construct a deep learner that finds the weights and bias terms which minimize

$$\hat{W}, \hat{b} = \underset{W,b}{\operatorname{argmin}} \frac{1}{T} \sum_{i=1}^{T} \mathcal{L}(Y^{(i)}, \hat{Y}(X^{(i)})) + \lambda \phi(W, b),$$

with negative cross-entropy loss, corresponding to multi-class logistic regression:

$$\mathcal{L}(Y, \hat{Y}) = -\sum_{k=1}^{K} Y_k \log \hat{Y}_k. \tag{6}$$

The exact feed forward architecture and weight matrix sizes of our deep learner are given by

$$\text{response}: \hat{Y}_k = \text{softmax}(Z^{L-1}) = \frac{\exp(Z_k^{L-1})}{\sum_{j=1}^{K} \exp(Z_j^{L-1})},$$

$$\text{hidden states}: Z^\ell = \max\left(W^\ell Z^{\ell-1} + b^\ell, 0\right), \ 1 \leq \ell < L,$$

where $L = 5$ and the network is tapered so that

$$W^1 \in \mathbb{R}^{300 \times 401}, W^2 \in \mathbb{R}^{200 \times 300}, W^3 \in \mathbb{R}^{100 \times 200}, W^4 \in \mathbb{R}^{50 \times 100} \text{ and } W^5 \in \mathbb{R}^{3 \times 50}.$$

**Alternative architectures** An alternative architecture is the RNN given in Section 2 that would be configured as

$$\text{response}: \hat{Y}_t = \text{softmax}(W^2 Z_t + b^2),$$

$$\text{hidden states}: Z_{t-j} = \tanh(W^1 [Z_{t-j-1}, X_{t-j}] + b^1), \ j \in \{k, \ldots, 0\}.$$

where $W_x^1 \in \mathbb{R}^{40 \times 11}, W_z^1 \in \mathbb{R}^{40 \times 40}, W_z^2 \in \mathbb{R}^{3 \times 40}$ and the hidden states are initialized to zero. Further details of the implementation and results using a RNN for this application are given in (Dixon, 2017).

**Training** We use the SGD method, implemented in `Python`'s `TensorFlow` (Abadi et al., 2016) framework, to find the optimal network weights, bias terms and regularization param-eters. We employ an exponentially decaying learning rate schedule with an initial value of $10^{-2}$. The optimal $\ell_2$ regularization is found, via a grid-search, to be $\lambda_2 = 0.01$. The Glorot and Bengio method is used to initialize the weights of the network (Glorot & Bengio, 2010).

Times series cross-validation is performed using a separate balanced training set and unbalanced validation and test sets, the latter two are each of size $2 \times 10^5$ observations. To avoid look ahead bias, each set represents a contiguous sampling period with the training set containing the earlier observations and the verification and test sets containing the most recent observations. The out-of-sample model performance on the verification set is used as the criteria for selecting our final deep learning architecture. Each experiment is run for 2500 epochs with a mini-batch size of 32 drawn from the balanced training set of 298,062 observations of 440 variables. These 440 variables are initially chosen from 10 liquidity imbalance ratios lagged up to 40 past observations and an additional lagged variable representing the relative size of the aggressors. Elastic-net ($\alpha = 0.5$), with a weight matrix $W^0 \in \mathbb{R}^{401 \times 440}$, is used for regularization and variable selection. The gridded search to find the optimal network architecture and regularization parameters takes several days on a graphics processing unit (GPU). The search yields several candidate architectures and parameter values.

Table 3 compares the performance of the deep learner with the elastic net method, implemented in the R package `glmnet` (Friedman, Hastie, & Tibshirani, 2010; Simon, Friedman, Hastie, & Tibshirani, 2011), for predicting the next price movement. The elastic-net method, with $\alpha = 0.5$, exhibits an out-of-sample classification accuracy of 49.6%. However, due to the imbalance of the data, we use the $F1$ score - the geometric mean of the precision and recall:

$$F1 = 2 \frac{\text{precision} \cdot \text{recall}}{\text{precision} + \text{recall}}.$$

The $F1$ score is designed for binary classification problems. When the data has more than two classes, the $F1$ score is provided for each class. The score is highest for the zero label corresponding to a prediction of a stationary mid-price over the next interval. The $F1$ scores for a predicted up-tick $F1(1)$ and down-tick $F1(-1)$ are also shown. The deep learners exhibit a higher accuracy of $81.7\%$ and higher $F1$ scores for each class.

Figure 10 compares the Receiver Operator Characteristic (ROC) curves for the deep learner and the elastic net method for (left) downward, (middle) neutral, or (right) upward price prediction. The plot is constructed by varying the probability threshold (a.k.a. cut-points) for positive classification over the interval $[0.5, 1)$ and estimating the true positive and true negative rate of each model. In each case, the deep learner is observed to out-perform the elastic

| Model | F1 (-1) | F1(0) | F1 (1) | Accuracy |
|---|---|---|---|---|
| Elastic-Net | 0.116 | 0.649 | 0.108 | 0.496 |
| DL with Elastic-Net | 0.201 | 0.897 | 0.186 | 0.817 |

Table 3: *The F1 scores and and classification accuracy are compared between the elastic net model and the combined deep learner and elastic net model. The deep learners exhibit a higher accuracy and higher F1 scores for each class.*

net method. The dashed line shows the performance of a white-noise classifier.

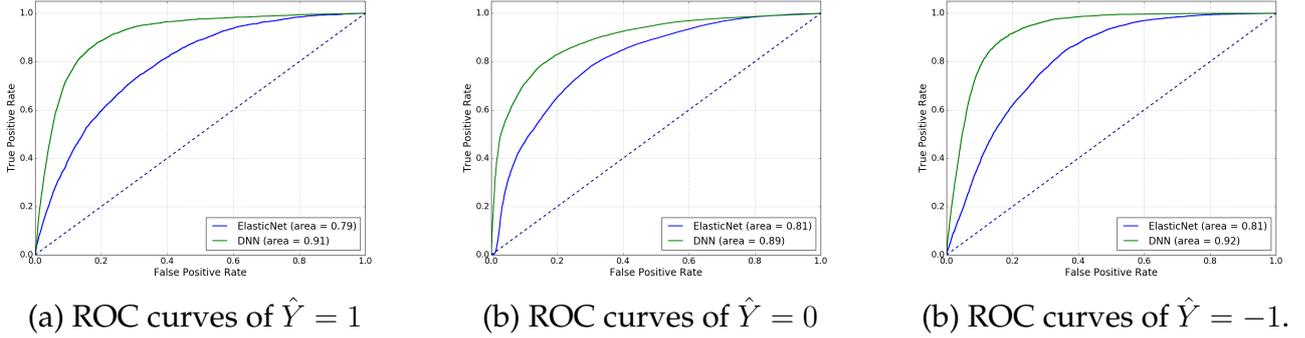

(a) ROC curves of $\hat{Y} = 1$  (b) ROC curves of $\hat{Y} = 0$  (b) ROC curves of $\hat{Y} = -1$.

Figure 10: The Receiver Operator Characteristic (ROC) curves of the deep learner and the elastic net method are shown for (left) downward, (middle) neutral, or (right) upward next price movement prediction.

Figure 11 shows the learning curves for each of the three labels corresponding to a downward, neutral, or upward price movement. These learning curves, showing the F1 score on the training and test set against the size of the training set, are used to assess the bias-variance tradeoff of the feature set. Each training set, of size shown by the x-axis, is sampled from the full training set of balanced observations. The model is trained on this subset and the F1 score of each label is measured in-sample and out-of-sample. The sampling is repeated to infer a distribution for each of the in-sample and out-of-sample F1 scores. The mean and confidence band of the F1 scores, at one standard deviation, are shown in each plot.

Using the learning curve, the size of the training set is chosen so that the variance, that is the difference between the F1 score of the classifier on the training set and test set, is sufficiently low. The variance is observed to reduce with an increased training set size and suggests that the model is not-overfitting. The bias on the test set is also observed to reduce with increased training set size.

Figure 12 compares the observed ESU6 mid-price movements with the deep learner forecasted price movements over one 1 milli-second intervals between 12:22:20 and 12:24:20 CST. The bottom three panels show the corresponding probabilities of predicting each class. A probability threshold of $0.65$ for the up-tick or down-tick classification is chosen here for illustrative purposes. Predicting the price movement protects high frequency market makers from adverse selection. False positives or negatives, when the observed mid-price is stationarity, oftentimes result in unnecessary order cancellations and loss of queue position. The false prediction of a stationary mid-price, or a false positive (negative) when the observed mid-price is negative (positive) lead to adverse selection.

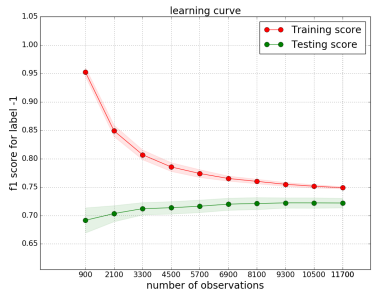 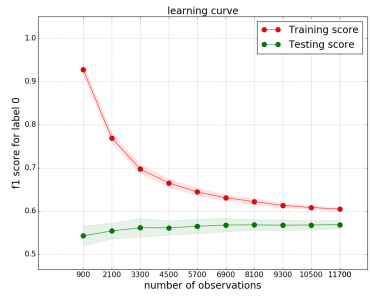 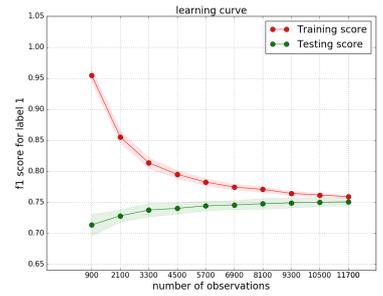

(a) DNN F1-score of $\hat{Y} = 1$   (b) DNN F1-score of $\hat{Y} = 0$   (b) DNN F1-score of $\hat{Y} = -1$.

Figure 11: *The learning curves of the deep learner are used to assess the bias-variance tradeoff and are shown for (left) downward, (middle) neutral, or (right) upward price prediction. The variance is observed to reduce with an increased training set size and shows that the deep learning is not-overfitting. The bias on the test set is also observed to reduce with increased training set size.*

## 5 Discussion

Deep learning architectures stand out from other machine learning methods for their ability to handle complex interactions and nonlinearities. By viewing a spatio-temporal dataset as 'image-like', we show the gains carry over to predicting sharp changes to spatial flow data in traffic and high frequency trading datasets.

Deep learning methods have some advantages and caveats. The key advantages are: (i) Modern software frameworks allow to easily implement deep learning architectures, (ii) More flexible learners compared to additive and tree models. The key caveats are (i) Model interpretability, (ii) It is time consuming to build models, some steps are ad-hoc and require modeler's attention, training times are longer compared to GLMs or Tree models, (iii) Due to the nesting of layers, statistical inference cannot always be applied to deep learning (Polson et al., 2017).

Yet, deep learning provides a very fruitful linear of research particularly in empirical asset pricing studies. Given the temporal nature, studying more complex architectures than recurrent and feed-forward neural networks, and using neural turing machines (NTMs) or long short-term memory (LSTMs) seems a very promising area for future statistical research. Finally, given the algorithmic nature of DL methods, understanding how they capture traditional physical models is also of interest. Much of the gains in other applied areas is the advantage of deep layers, see for example (Montúfar, Pascanu, Cho, & Bengio, 2014). Our work shows that this carries over to spatio-temporal modeling.

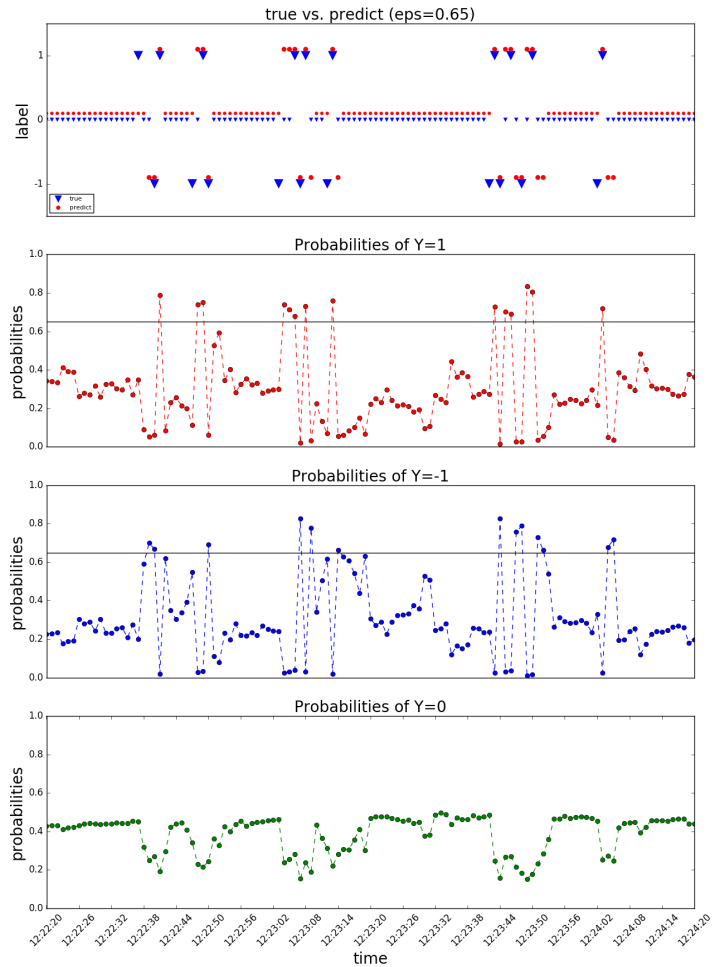

Figure 12: *The (top) comparison of the observed ESU6 mid-price movements with the deep learner forecasted price movements over one milli-second intervals between 12:22:20 and 12:24:20 CST. The bottom three panels show the corresponding probabilities of predicting each class. A probability threshold of* $0.65$ *for the up-tick or down-tick classification is chosen here for illustrative purposes.*

# Appendix A

## 5.1 Training, Validation and Testing

Deep learning is a data-driven approach which focuses on finding structure in large data sets. The main tools for variable or predictor selection are regularization and dropout. Out-of-sample predictive performance helps assess the optimal amount of regularization, the problem of finding the optimal hyper-parameter selection. There is still a very Bayesian flavor to the modeling procedure and the researcher follows two key steps:

1. Training phase: pair the input with expected output, until a sufficiently close match has been found. Gauss' original least squares procedure is a common example.

2. Validation and test phase: assess how well the deep learner has been trained for out-of-sample prediction. This depends on the size of your data, the value you would like to predict, the input, etc., and various model properties including the mean-error for numeric predictors and classification-errors for classifiers.

Often, the validation phase is split into two parts.

2.a First, estimate the out-of-sample accuracy of all approaches (a.k.a. validation).

2.b Second, compare the models and select the best performing approach based on the validation data (a.k.a. verification).

Step 2.b. can be skipped if there is no need to select an appropriate model from several rivaling approaches. The researcher then only needs to partition the data set into a training and test set.

To construct and evaluate a learning machine, we start with training data of input-output pairs $D = \{Y^{(i)}, X^{(i)}\}_{i=1}^T$. The goal is to find the machine learner $Y = F(X)$, where we have a loss function $\mathcal{L}(Y, \hat{Y})$ for a predictor, $\hat{Y}$, of the output signal, $Y$. In many cases, there's an underlying probability model, $p(Y \mid \hat{Y})$, then the loss function is the negative log probability $\mathcal{L}(Y, \hat{Y}) = -\log p(Y \mid \hat{Y})$. For example, under a Gaussian model $\mathcal{L}(Y, \hat{Y}) = ||Y - \hat{Y}||^2$ is a $L^2$ norm, for binary classification, $\mathcal{L}(Y, \hat{Y}) = -Y \log \hat{Y}$ is the negative cross-entropy. In its

simplest form, we then solve an optimization problem

$$\underset{W,b}{\text{minimize}} f(W,b) + \lambda \phi(W,b)$$

$$f(W,b) = \frac{1}{T} \sum_{i=1}^{T} \mathcal{L}(Y^{(i)}, \hat{Y}(X^{(i)}))$$

with a regularization penalty, $\phi(W,b)$. Here $\lambda$ is a global regularization parameter which we tune using the out-of-sample predictive mean-squared error (MSE) of the model. The regularization penalty, $\phi(W,b)$, introduces a bias-variance tradeoff. $\nabla \mathcal{L}$ is given in closed form by a chain rule and, through back-propagation, each layer's weights $\hat{W}^l$ are fitted with stochastic gradient descent.

## 5.2 Stochastic gradient descent (SGD)

Stochastic gradient descent (SGD) method or its variation is typically used to find the deep learning model weights by minimizing the penalized loss function, $f(W,b)$. The method minimizes the function by taking a negative step along an estimate $g^k$ of the gradient $\nabla f(W^k, b^k)$ at iteration $k$. The approximate gradient is calculated by

$$g^k = \frac{1}{b_k} \sum_{i \in E_k} \nabla \mathcal{L}_{W,b}(Y^{(i)}, \hat{Y}^k(X^{(i)})),$$

where $E_k \subset \{1, \ldots, T\}$ and $b_k = |E_k|$ is the number of elements in $E_k$ (a.k.a. batch size). When $b_k > 1$ the algorithm is called batch SGD and simply SGD otherwise. A usual strategy to choose subset $E$ is to go cyclically and pick consecutive elements of $\{1, \ldots, T\}$ and $E_{k+1} = [E_k \mod T] + 1$, where modular arithmetic is applied to the set. The approximated direction $g^k$ is calculated using a chain rule (a.k.a. back-propagation) for deep learning. It is an unbiased estimator of $\nabla f(W^k, b^k)$, and we have

$$\mathrm{E}(g^k) = \frac{1}{T} \sum_{i=1}^{T} \nabla \mathcal{L}_{W,b} \left( Y^{(i)}, \hat{Y}^k(X^{(i)}) \right) = \nabla f(W^k, b^k).$$

At each iteration, we update the solution $(W,b)^{k+1} = (W,b)^k - t_k g^k$.

Deep learning applications use a step size $t_k$ (a.k.a. learning rate) as constant or a reduction strategy of the form, $t_k = a \exp(-kt)$. Appropriate learning rates or the hyper-parameters of reduction schedule are usually found empirically from numerical experiments and observa-tions of the loss function progression.

One disadvantage of SGD is that the descent in $f$ is not guaranteed or can be very slow at every iteration. Furthermore, the variance of the gradient estimate $g^k$ is near zero as the iterates converge to a solution. To tackle those problems a coordinate descent (CD) and momentum-based modifications of SGD are used. Each CD step evaluates a single component $E_k$ of the gradient $\nabla f$ at the current point and then updates the $E_k$th component of the variable vector in the negative gradient direction. The momentum-based versions of SGD or so-called accelerated algorithms were originally proposed by (Nesterov, 2013).

The use of momentum in the choice of step in the search direction combines new gradient information with the previous search direction. These methods are also related to other classical techniques such as the heavy-ball method and conjugate gradient methods. Empirically momentum-based methods show a far better convergence for deep learning networks. The key idea is that the gradient only influences changes in the "velocity" of the update

$$v^{k+1} = \mu v^k - t_k g^k,$$
$$(W,b)^{k+1} = (W,b)^k + v^k.$$

The parameter $\mu$ controls the dumping effect on the rate of update of the variables. The physical analogy is the reduction in kinetic energy that allows "slow down" the movements at the minima. This parameter is also chosen empirically using cross-validation.

Nesterov's momentum method (a.k.a. Nesterov acceleration) instead calculate gradient at the point predicted by the momentum. We can think of it as a look-ahead strategy. The resulting update equations are

$$v^{k+1} = \mu v^k - t_k g((W,b)^k + v^k),$$
$$(W,b)^{k+1} = (W,b)^k + v^k.$$

Another popular modification to the SGD method is the AdaGrad method, which adaptively scales each of the learning parameters at each iteration

$$c^{k+1} = c^k + g((W,b)^k)^2,$$
$$(W,b)^{k+1} = (W,b)^k - t_k g(W,b)^k)/(\sqrt{c^{k+1}} - a),$$

where $a$ is usually a small number, e.g. $a = 10^{-6}$ that prevents dividing by zero. PRM-

Sprop takes the AdaGrad idea further and places more weight on recent values of the gradient squared to scale the update direction, i.e. we have

$$c^{k+1} = dc^k + (1-d)g((W,b)^k)^2.$$

The Adam method combines both PRMSprop and momentum methods and leads to the following update equations

$$v^{k+1} = \mu v^k - (1-\mu)t_k g((W,b)^k + v^k),$$
$$c^{k+1} = dc^k + (1-d)g((W,b)^k)^2,$$
$$(W,b)^{k+1} = (W,b)^k - t_k v^{k+1}/(\sqrt{c^{k+1}} - a).$$

Second order methods solve the optimization problem by solving a system of nonlinear equations $\nabla f(W,b) = 0$ with the Newton's method

$$(W,b)^+ = (W,b) - \{\nabla^2 f(W,b)\}^{-1}\nabla f(W,b).$$

SGD simply approximates $\nabla^2 f(W,b)$ by $1/t$. The advantages of a second order method include much faster convergence rates and insensitivity to the conditioning of the problem. In practice, second order methods are rarely used for deep learning applications (Dean et al., 2012). The major disadvantage is the inability to train the model using batches of data as SGD does. Since typical deep learning models relies on large scale data sets, second order methods become memory and computationally prohibitive at even modest-sized training data sets.

However batching alone is not sufficient to scale SGD methods to large-scale problems on modern high performance computers. Back-propagation through a chain-rule creates an inherit sequential dependency in the weight updates which limits the dataset dimensions for the deep learner. (Polson, Willard, & Heidari, 2015) consider a proximal Newton method, a Bayesian optimization technique which provides an efficient solution for estimation and optimization of such models and for calculating a regularization path. The authors present a splitting approach, alternating direction method of multipliers (ADMM), which overcomes the inherent bottle-necks in back-propagation by providing a simultaneous block update of parameters at all layers. ADMM facilitates the use of large-scale computing.

A significant factor in the widespread adoption of deep learning has been the creation of `TensorFlow` (Abadi et al., 2016), an interface for easily expressing machine learning al-gorithms and mapping compute intensive operations onto a wide variety of different hardware platforms and in particular GPU cards. Recently, `TensorFlow` has been augmented by `Edward` (Tran et al., 2017) to combine concepts in Bayesian statistics and probabilistic programming with deep learning.

## 5.3 Model Averaging via Dropout

Dropout is a computationally efficient technique to reduce model variance by considering many model configurations and then averaging the predictions. The input space $X = (X_1, \ldots, X_p)$, where $p$ is large, needs dimension reduction techniques which are designed to avoid overfitting in the training process. Dropout works by removing input dimensions in $X$ randomly with a given probability $\theta$. The probability, $\theta$, can be viewed as a further hyper-parameter (like $\lambda$) which can be tuned via cross-validation. Heuristically, if there are 1000 variables, then a choice of $\theta = 0.1$ will result in a search for models with 100 variables. The dropout architecture with stochastic search for the predictors can be used

$$\begin{aligned}
D_i^l &\sim \text{Ber}(\theta), \\
\tilde{Z}^l &= D^l \star Z^l,\ 1 \leq l < L, \\
Z^l &= f^l(W^l \tilde{Z}^{l-1} + b^l).
\end{aligned}$$

Effectively, this replaces the input $X$ by $D \star X$, where $\star$ denotes the element-wise product and $D$ is a vector of independent Bernoulli Ber($\theta$) distributed random variables. The overall objective function is closely related to ridge regression with a g-prior (Heaton, Polson, & Witte, 2017).